\documentclass[conference]{IEEEtran}
\IEEEoverridecommandlockouts

\usepackage{cite}
\usepackage{amsmath,amssymb,amsfonts}
\usepackage{graphicx}
\usepackage{booktabs}
\usepackage{xcolor}
\usepackage{url}
\usepackage[hidelinks]{hyperref}

\newenvironment{algobox}[1]
  {\par\noindent\rule{\linewidth}{0.8pt}\par\textbf{#1}\par
   \vspace{-0.3em}\rule{\linewidth}{0.3pt}\par\small\begin{enumerate}
   \setlength\itemsep{0pt}\setlength\parskip{0pt}}
  {\end{enumerate}\vspace{-0.2em}\rule{\linewidth}{0.8pt}\par}

\newcommand{\R}{\mathbb{R}}

\begin{document}

\title{MST-Direct at Scale: Multivariate and Conditional\\
Geostatistical Simulation via Sinkhorn Optimal Transport%
\thanks{Follow-up to the author's prior work (arXiv:2603.18036), which
introduced MST-Direct in the bivariate, unconditional, small-grid setting; this
paper contributes the scalable, multivariate, and conditional extensions.}}

\author{\IEEEauthorblockN{Tcharlies Bachmann Schmitz}
\IEEEauthorblockA{Data Science, PX.Center\\
Joinville, Brasil \\
tcharliesschmitz@gmail.com \quad ORCID: 0009-0007-5467-1327}}

\maketitle

\begin{abstract}
This paper extends MST-Direct, the Matching-via-Sinkhorn-Transport approach to
multivariate geostatistical simulation, from the bivariate, unconditional,
small-grid setting of the original formulation to the \emph{multivariate},
\emph{conditional}, and \emph{large-grid} regime. We address the three open
limitations identified in that work: (i) scalability beyond a few thousand
nodes, through a sparse, candidate-restricted Sinkhorn matcher with
$O(n\,C)$ memory; (ii) extension to many variables, by matching the target
value tuples onto an independent FFT-MA Gaussian backbone that carries a
prescribed variogram; and (iii) hard-data conditioning, by pinning the data
tuples to their locations while conditioning the backbone by kriging. Because
the transport plan remains a permutation of the target tuples, the
multivariate joint distribution is preserved \emph{exactly}. We validate the
method on the same 6-variate, heteroscedastic, strongly non-linear reference
distribution used by Direct Multivariate Simulation (DMS), in both
unconditional ($200\times200$) and conditional ($100\times100$, 200 hard data)
settings, and we benchmark it against the Projection Pursuit Multivariate
Transform (PPMT). MST-Direct reproduces the joint distribution with zero
histogram error, honours the hard data exactly, and reproduces the prescribed
spatial correlation, whereas PPMT remains an approximation.
\end{abstract}


\section{Introduction}
Geostatistical simulation generates multiple realizations of geological models
that replicate the spatial features of the data and support uncertainty
quantification \cite{Goovaerts1997,ChilesDelfiner1999}. Many geological
relationships are strongly non-linear and heteroscedastic and cannot be
characterized by the linear correlation coefficient on which classical
multivariate methods rely. The Stepwise Conditional Transformation (SCT)
\cite{Leuangthong2003sct}, the Projection Pursuit Multivariate Transform (PPMT)
\cite{Barnett2014ppmt}, and Direct Multivariate Simulation (DMS)
\cite{deFigueiredo2021dms} address this by transforming the variables to a
Gaussian space and back; the transform is exact only asymptotically and its
error grows with the number of variables.

In previous work we introduced \emph{MST-Direct} (Matching via Sinkhorn
Transport) \cite{Schmitz2026mst}, which casts simulation as an
entropy-regularized optimal-transport (OT) problem
\cite{Villani2009ot,PeyreCuturi2019,Cuturi2013sinkhorn}: the target value tuples
are \emph{matched} onto a spatial template through the Sinkhorn algorithm
\cite{Sinkhorn1964} augmented with a relational $k$-nearest-neighbour term, and
the resulting permutation reproduces the spatial structure while preserving the
joint distribution exactly. That formulation achieved $100\%$ shape
preservation on five complex \emph{bivariate} relationships, but was
demonstrated only on a $25\times25$ grid against Gaussian-copula and
LU-decomposition baselines, and explicitly left three problems open
\cite{Schmitz2026mst}: scalability beyond $\sim$10\,000 nodes, extension to
more than two variables, and conditional simulation with hard data.

This paper resolves those three limitations and benchmarks the method on the
DMS validation problem. Our contributions are:
\begin{enumerate}
\item a \emph{scalable} MST-Direct matcher --- sparse, candidate-restricted
Sinkhorn with greedy bijection completion --- that runs the $200\times200$
($40\,000$-node) and $100\times100$ grids in well under a minute;
\item a \emph{multivariate} formulation that matches the target cloud onto an
independent FFT-MA Gaussian backbone \cite{LeRavalec2000fftma} carrying the
prescribed variogram, demonstrated on six variables;
\item \emph{conditional} simulation that honours hard data exactly, by pinning
the data tuples and conditioning the backbone by simple kriging;
\item a head-to-head validation against PPMT on the same 6-variate,
heteroscedastic, non-linear reference distribution as DMS
\cite{deFigueiredo2021dms}.
\end{enumerate}

\section{Method}
\label{sec:method}

We first summarize the MST-Direct principle introduced in
\cite{Schmitz2026mst}; the rest of this section presents the three extensions
that constitute the present contribution (scalable matching, the multivariate
backbone, and conditioning). Let $\{\mathbf{z}^{(k)}\}_{k=1}^{N}$,
$\mathbf{z}^{(k)}\in\R^{d}$, be a sample of $N$ tuples from a target $d$-variate
non-parametric distribution $p(z_1,\dots,z_d)$ (the data, or a representative
training set), to be placed on a grid of $n=N$ locations so that the realization
reproduces both $p$ and a prescribed spatial covariance. Writing the target
tuples and a spatial template $\{\mathbf{g}_j\}$ as discrete measures, a
realization is a coupling that assigns one tuple to each location; MST-Direct
selects it by the entropy-regularized optimal-transport problem
\cite{Cuturi2013sinkhorn}
\begin{equation}
\min_{M\in\Pi}\;\langle C,M\rangle-\tfrac1\beta H(M),
\label{eq:ot}
\end{equation}
over the transport polytope $\Pi=\{M\!\geq\!0:\,M\mathbf 1=\tfrac1n\mathbf 1,\,
M^{\!\top}\mathbf 1=\tfrac1n\mathbf 1\}$, with
$C_{ij}=\lVert\tilde{\mathbf g}_i-\tilde{\mathbf z}^{(j)}\rVert^2$ the
squared distance in the per-variable standardized space, $H$ the entropy, and
$\beta$ the regularization. The solution
$M=\mathrm{diag}(\mathbf r)K\,\mathrm{diag}(\mathbf c)$, $K=\exp(-\beta C)$, is
found by the Sinkhorn fixed point \cite{Sinkhorn1964}, run in the log domain.

A relational $k$-nearest-neighbour reward then augments the cost so that
spatially adjacent locations are matched to mutually similar tuples
\cite{Schmitz2026mst}; with $A$ the row-normalized grid adjacency,
\begin{equation}
M_{ij}\propto\exp\!\big(\beta(-C_{ij}+\lambda\,[A M A^{\!\top}]_{ij})\big),
\label{eq:rel}
\end{equation}
solved by alternating the reward $AMA^{\!\top}$ with Sinkhorn normalization. The
soft coupling is rounded to a permutation $\pi$ by greedy assignment in order of
decreasing confidence. Since $\pi$ is a bijection onto the target tuples, the
realization $\mathbf z^{*}_i=\mathbf z^{(\pi(i))}$ is the \emph{same multiset} of
tuples: every marginal and every non-linear cross-dependence is reproduced
exactly. OT decides only \emph{where} each tuple goes.

The template carries the prescribed spatial structure. As in DMS we generate
$d$ \emph{independent} standard-Gaussian fields with the target (spherical)
variogram by FFT-MA \cite{LeRavalec2000fftma},
$\mathbf g^{(c)}=\mathcal F^{-1}\{\sqrt{S(\boldsymbol\omega)}\,\mathcal
F\{W^{(c)}\}\}$. Neighbouring locations receive similar feature vectors and,
after matching, similar tuples, transferring the variogram to every variable.
The matching itself imposes \emph{no} isotropy or stationarity requirement: any
covariance admissible by the chosen Gaussian simulator --- including anisotropic
models --- can be used in the backbone; we adopt an isotropic spherical model
only for comparability with DMS \cite{deFigueiredo2021dms}.

A dense coupling is $O(n^2)$ in memory and $\sim O(n^3)$ per relational
iteration, infeasible at $n=40\,000$ \cite{Schmitz2026mst}. We instead restrict
each location to its $C$ nearest target tuples (via a $k$-d tree), so the
coupling has $O(nC)$ nonzeros and the Sinkhorn updates become segment
reductions over the candidate edge list. Greedy rounding yields the
permutation; the few locations left without a free candidate are repaired by
nearest-unused assignment, guaranteeing a complete bijection. The relational
term \eqref{eq:rel} is applied as a few refinement passes pulling each query
toward $\overline{Z}_{\mathcal N(i)}$, the mean tuple currently assigned to the
spatial $k$-nearest neighbours $\mathcal N(i)$ of location $i$ (Algorithm~1).

\begin{algobox}{Algorithm 1: Scalable MST-Direct}
\item \textbf{input} target tuples $Z\in\R^{n\times d}$; coordinates; variogram;
hard data (optional).
\item standardize $Z$; build backbone $G$ (kriging-conditioned if hard data).
\item pin hard-data locations to their tuples; exclude from the pool.
\item candidates: $C$ nearest rows of $Z$ to each $G_i$ ($k$-d tree).
\item sparse log-domain Sinkhorn \eqref{eq:ot} on the candidate edges.
\item $r$ relational passes \eqref{eq:rel}:
$G_i\leftarrow(1-\lambda)G_i+\lambda\,\overline{Z}_{\mathcal N(i)}$, rematch.
\item greedy-round to a permutation $\pi$; repair leftovers.
\item \textbf{return} $\mathbf z^{*}_i=Z_{\pi(i)}$ (a permutation of $Z$).
\end{algobox}

Hard data $\{\mathbf z^{\mathrm d}_\ell\}$ at locations $\{\ell\}$ are honoured in
two steps. First, the backbone is conditioned to the data (in the matching
space) by simple kriging \cite{LeRavalec2000fftma}, so it interpolates the data
and keeps the prescribed covariance elsewhere. Second, the data locations are
\emph{pinned}: their tuples are fixed and removed from the pool, and only the
remaining locations enter \eqref{eq:ot}. The realization thus honours the data
exactly while the kriged backbone keeps the surroundings coherent.

\section{Experimental Design}
We reproduce the two experiments of DMS \cite{deFigueiredo2021dms} using
\emph{the same} 6-variate target distribution, which is strongly non-linear and
heteroscedastic (left panels of Figs.~\ref{fig:uncond} and \ref{fig:cond}). In
every comparison the target (\emph{real}) distribution is shown with the PPMT
and the MST-Direct realizations, in that order; diagonals are marginals and
off-diagonals bivariate histograms. We quantify the reproduction, for both
methods, by the mean squared error (MSE) between experimental and reference
histograms, and we assess the spatial reproduction by the experimental
variograms. We benchmark against PPMT \cite{Barnett2014ppmt} rather than the
Gaussian-copula or LU-decomposition baselines used in the bivariate study
\cite{Schmitz2026mst}: those baselines are Gaussian by construction and
therefore cannot reproduce complex non-linear joints, which would make the
comparison uninformative, whereas PPMT is a non-parametric multivariate
transform designed precisely for such distributions. PPMT is implemented with
projection-pursuit Gaussianization and uses the same FFT-MA backbone. The
unconditional grid is
$200\times200$ with an isotropic spherical variogram of range~40; the
conditional grid is $100\times100$ with 200 uniformly-located hard data and a
spherical variogram fitted from the data (range~$\approx16$, vs.\ 18 in
\cite{deFigueiredo2021dms}).

\section{Results}

\begin{figure*}[t]
\centering
\includegraphics[width=0.32\textwidth]{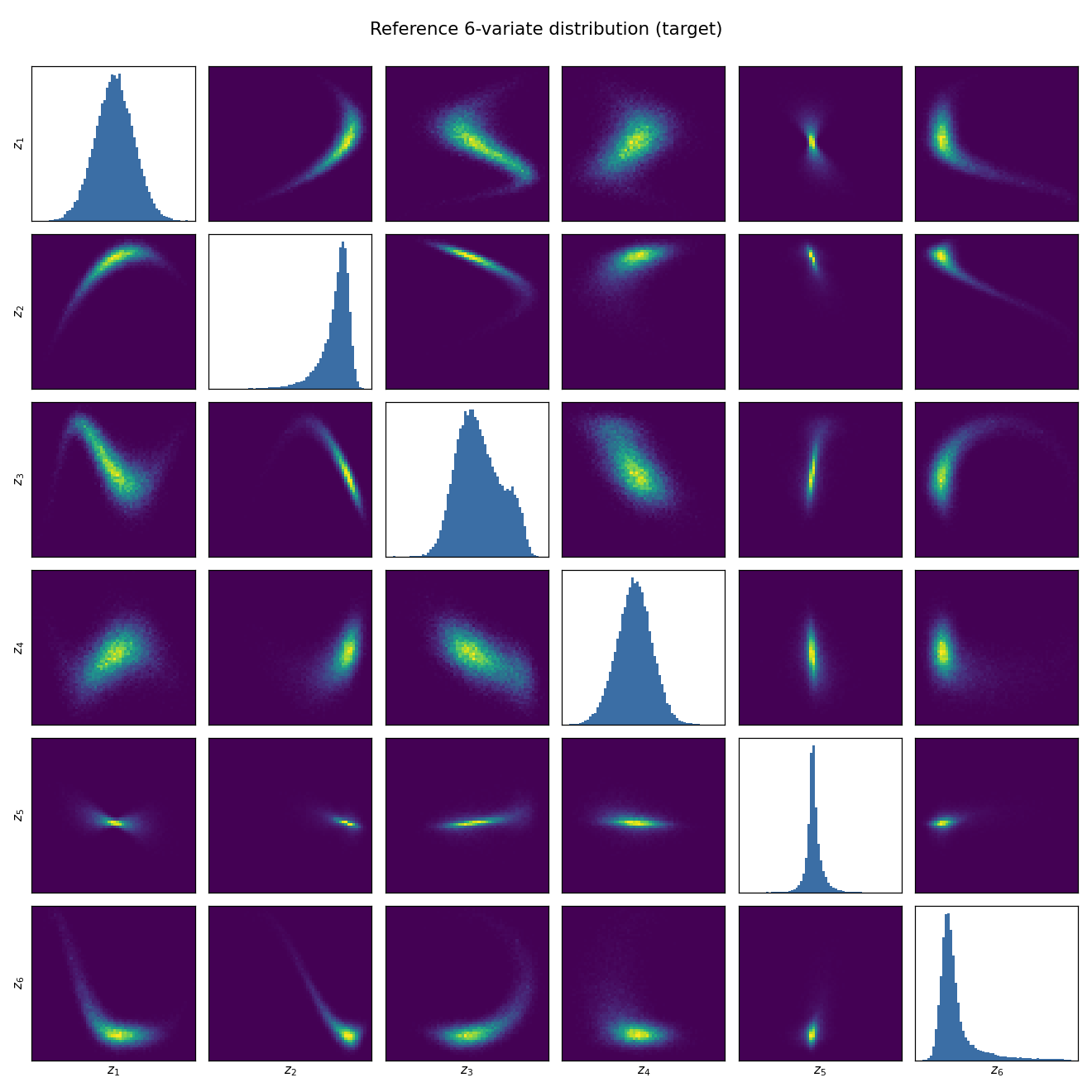}\hfill
\includegraphics[width=0.32\textwidth]{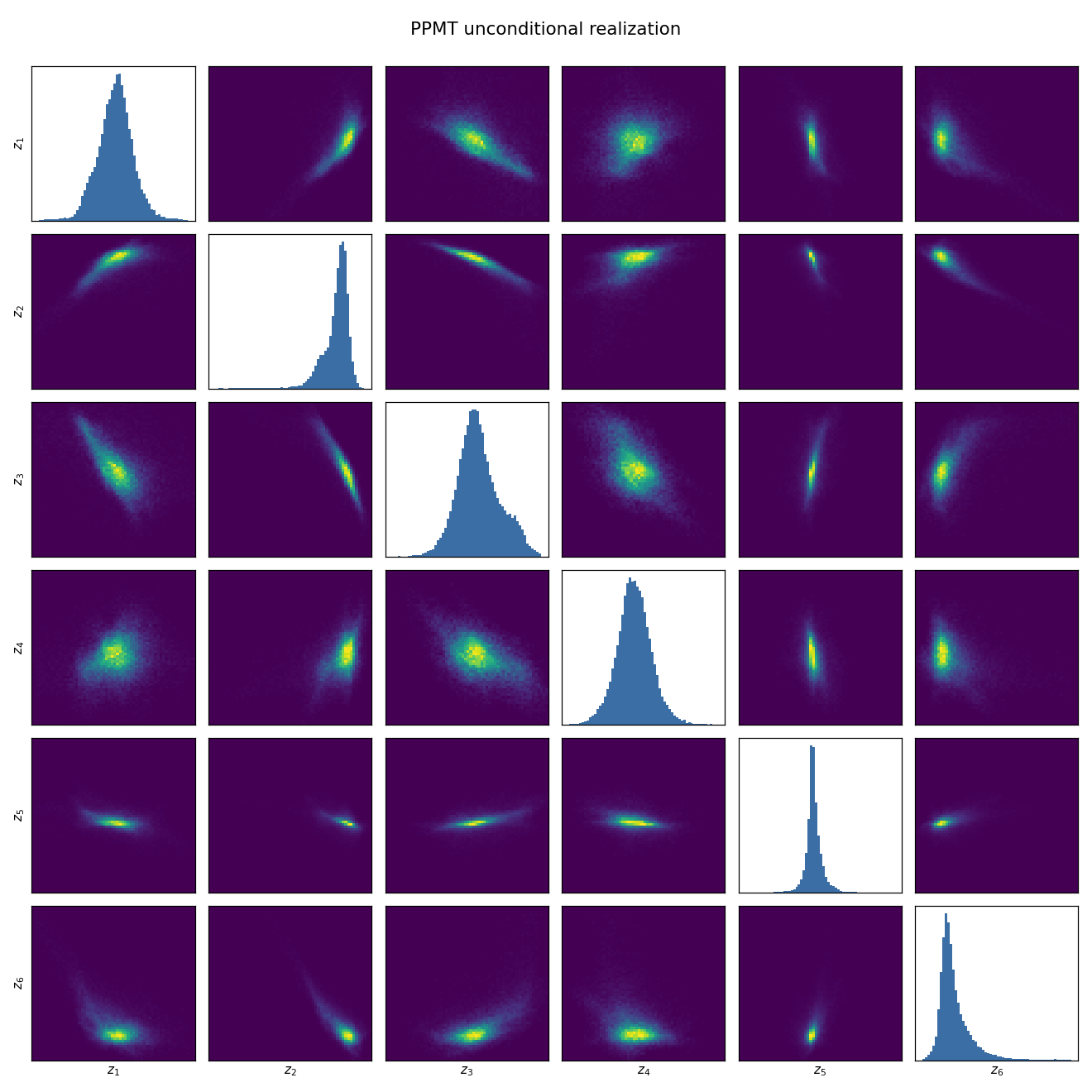}\hfill
\includegraphics[width=0.32\textwidth]{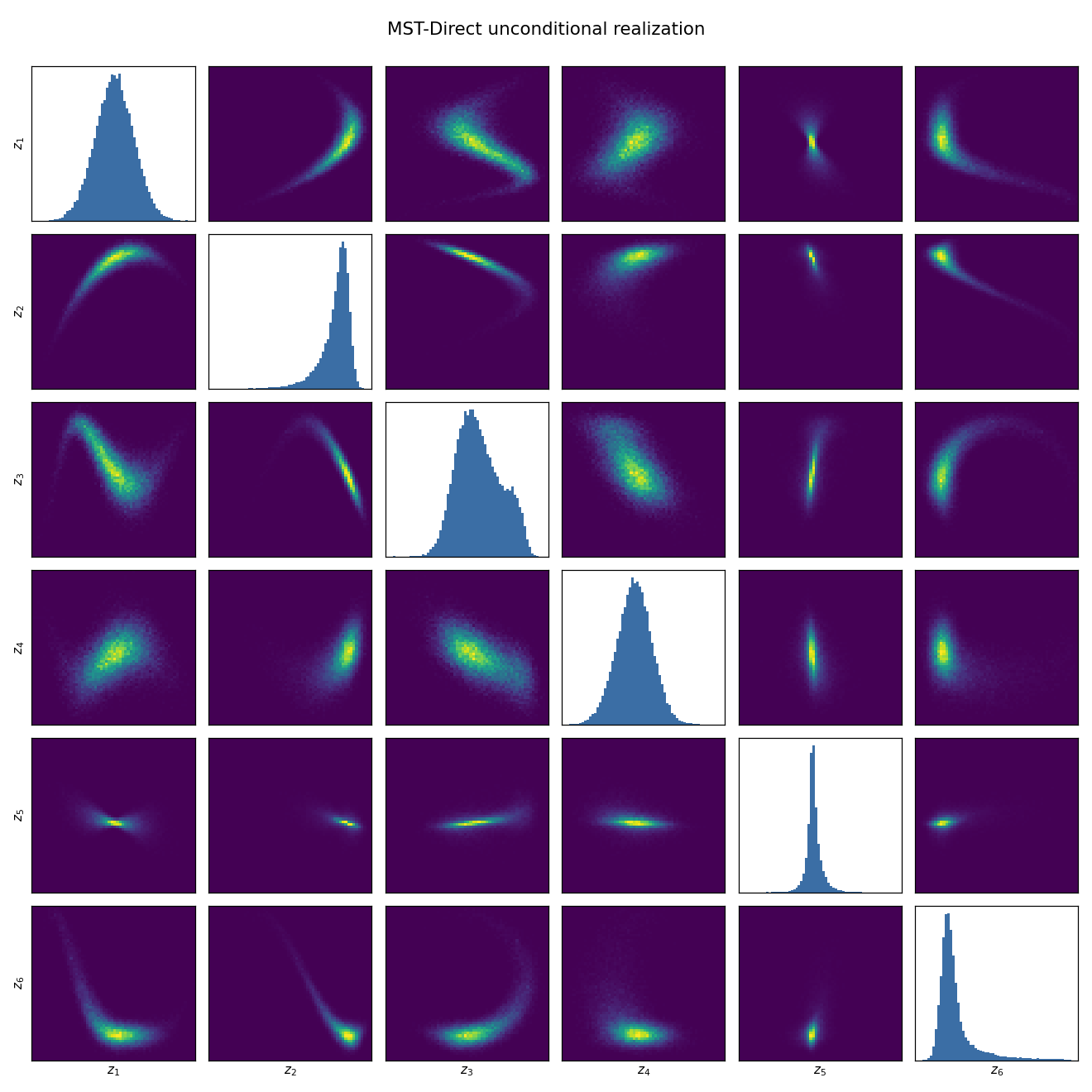}
\caption{Unconditional experimental distributions: \emph{real} (left), PPMT
(centre), MST-Direct (right). MST-Direct is identical to the reference; PPMT
distorts the fine non-linear joints.}
\label{fig:uncond}
\end{figure*}

\begin{figure*}[t]
\centering
\includegraphics[width=0.86\textwidth]{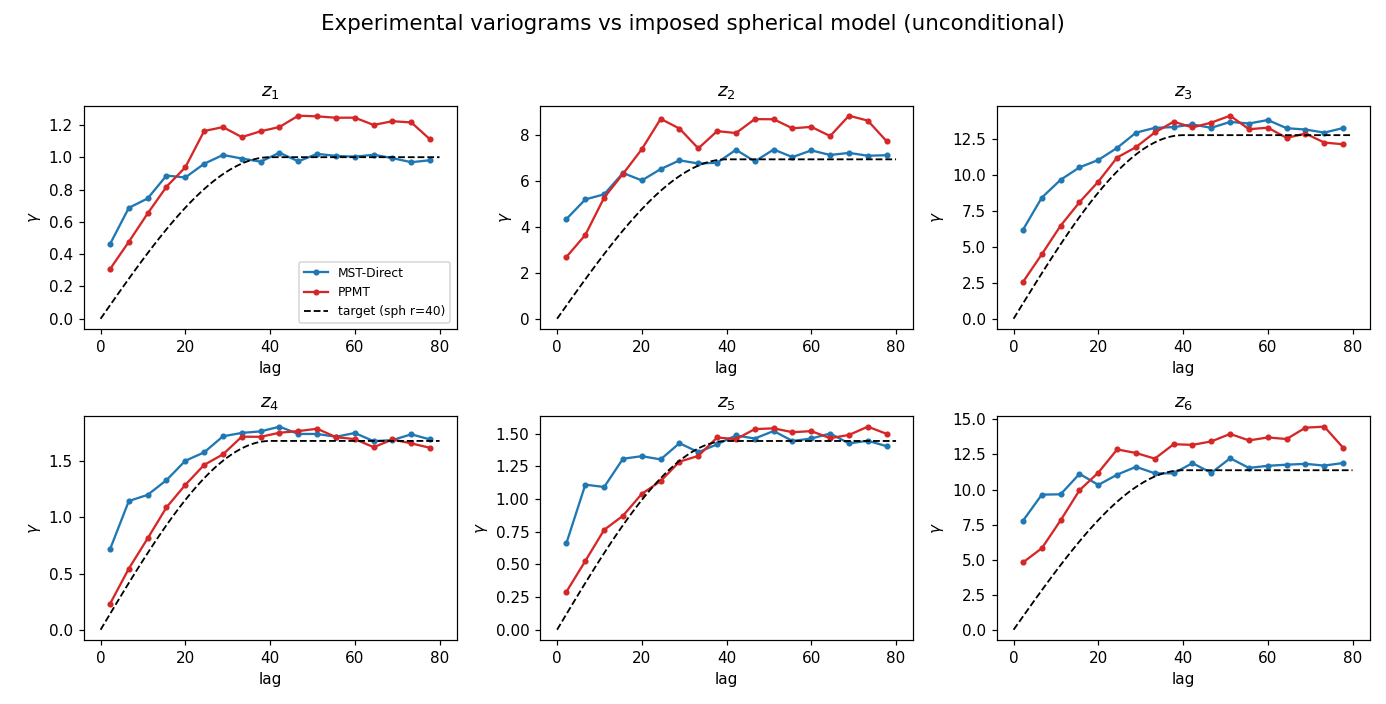}
\caption{Unconditional experimental variograms of MST-Direct (blue) and PPMT
(red) against the imposed spherical model with range 40 (dashed).}
\label{fig:uncondvario}
\end{figure*}

For the unconditional experiment ($200\times200$ grid, spherical range 40),
Fig.~\ref{fig:uncond} compares the experimental distributions. The MST-Direct
realization (right) is indistinguishable from the reference (left) --- a spatial
permutation of the same tuples --- whereas PPMT (centre) recovers the gross
structure but distorts the fine non-linear joints. Table~\ref{tab:uncond}
reports the histogram MSE for both methods: identically zero for MST-Direct and
of order $10^{-6}$--$10^{-5}$ for PPMT. Fig.~\ref{fig:uncondvario} shows that
both methods reproduce the imposed range-40 model; MST-Direct matches the sill
exactly (its variance equals the reference by construction), while PPMT
overshoots it by $10$--$20\%$. The scalable bijection leaves a mild short-lag
component, reduced by the relational passes.

\begin{table*}[t]
\centering
\caption{Unconditional histogram MSE vs.\ the reference ($\times10^{-5}$):
MST-Direct (left) and PPMT (right). Diagonal: marginals; off-diagonal:
bivariate histograms.}
\label{tab:uncond}
\footnotesize
\begin{tabular}{lcccccc}
\multicolumn{7}{c}{\textbf{MST-Direct}}\\
\toprule
 & $z_1$ & $z_2$ & $z_3$ & $z_4$ & $z_5$ & $z_6$\\
\midrule
$z_1$ & 0 & 0 & 0 & 0 & 0 & 0\\
$z_2$ & 0 & 0 & 0 & 0 & 0 & 0\\
$z_3$ & 0 & 0 & 0 & 0 & 0 & 0\\
$z_4$ & 0 & 0 & 0 & 0 & 0 & 0\\
$z_5$ & 0 & 0 & 0 & 0 & 0 & 0\\
$z_6$ & 0 & 0 & 0 & 0 & 0 & 0\\
\bottomrule
\end{tabular}
\hspace{2em}
\begin{tabular}{lcccccc}
\multicolumn{7}{c}{\textbf{PPMT}}\\
\toprule
 & $z_1$ & $z_2$ & $z_3$ & $z_4$ & $z_5$ & $z_6$\\
\midrule
$z_1$ & 2.71 & 0.02 & 0.01 & 0.01 & 0.04 & 0.02\\
$z_2$ & 0.02 & 2.49 & 0.02 & 0.01 & 0.02 & 0.02\\
$z_3$ & 0.01 & 0.02 & 2.62 & 0.01 & 0.02 & 0.01\\
$z_4$ & 0.01 & 0.01 & 0.01 & 0.82 & 0.01 & 0.01\\
$z_5$ & 0.04 & 0.02 & 0.02 & 0.01 & 0.90 & 0.02\\
$z_6$ & 0.02 & 0.02 & 0.01 & 0.01 & 0.02 & 2.00\\
\bottomrule
\end{tabular}
\end{table*}

\begin{figure*}[t]
\centering
\includegraphics[width=0.32\textwidth]{figures/reference_scatter.png}\hfill
\includegraphics[width=0.32\textwidth]{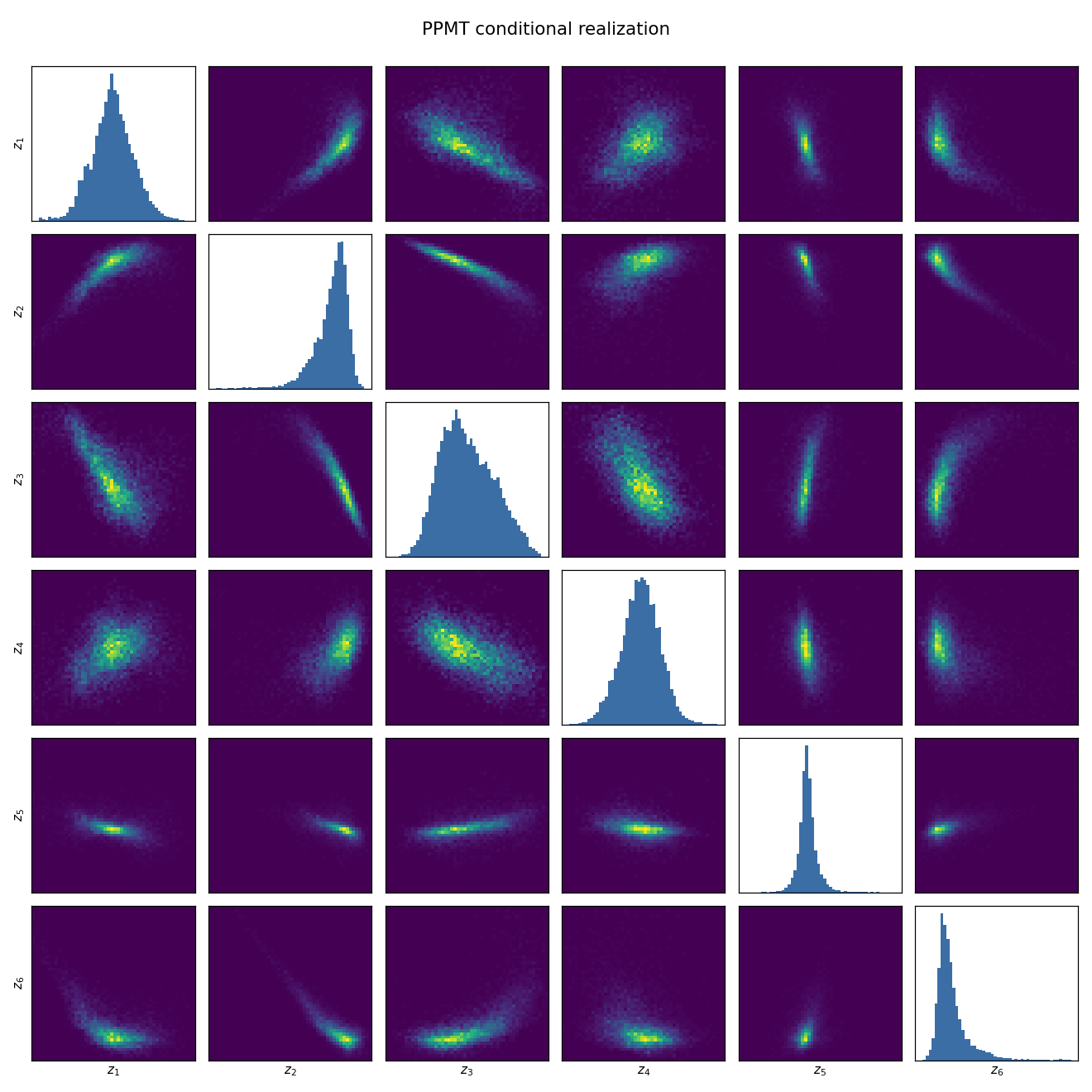}\hfill
\includegraphics[width=0.32\textwidth]{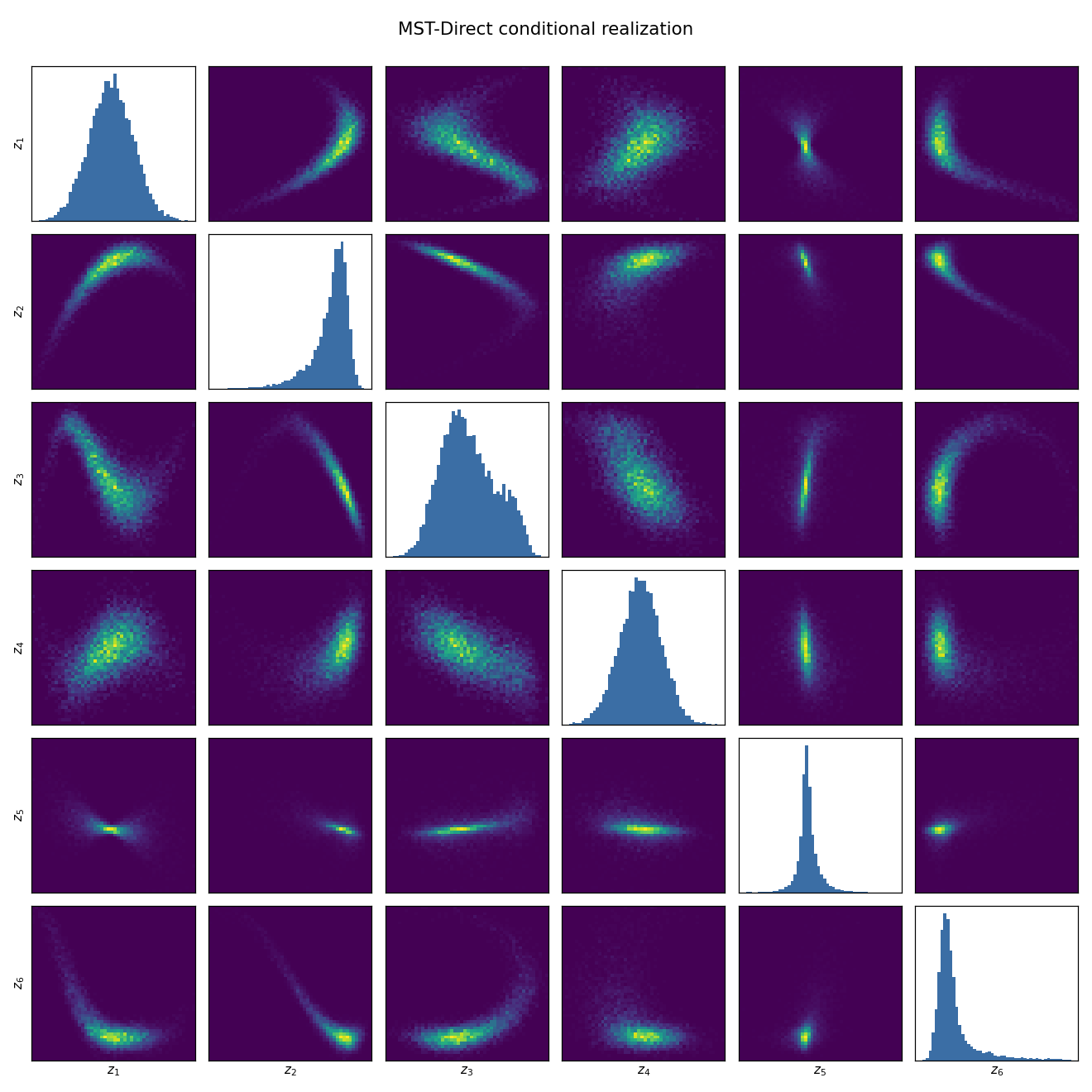}
\caption{Conditional experimental distributions: \emph{real} (left), PPMT
(centre), MST-Direct (right).}
\label{fig:cond}
\end{figure*}

\begin{figure*}[t]
\centering
\includegraphics[width=0.96\textwidth]{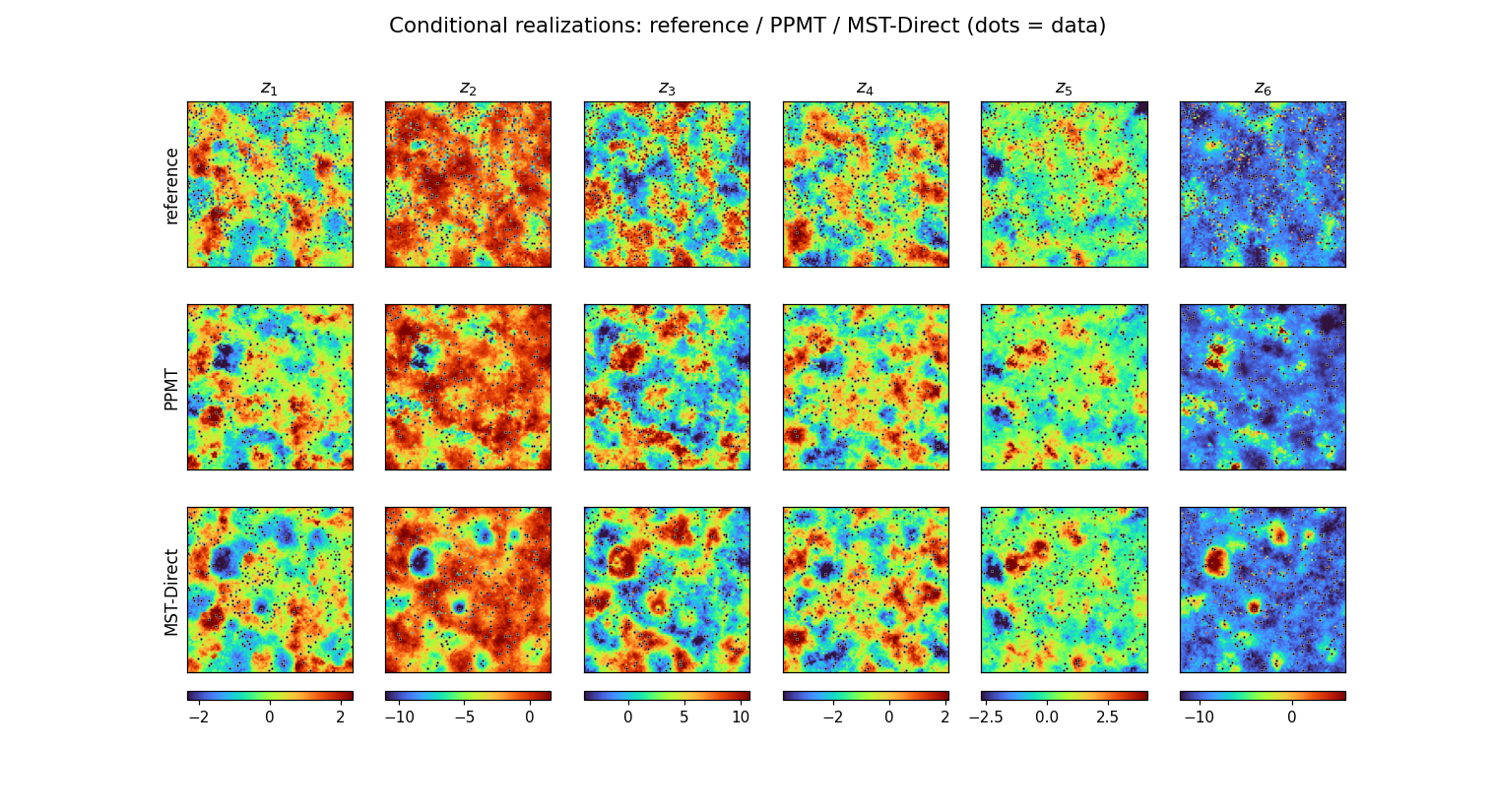}
\caption{Conditional realizations on the $100\times100$ grid, rows
\emph{reference / PPMT / MST-Direct}, columns $z_1$--$z_6$ (shared colour scale
per column). Black dots mark the 200 hard-data locations, honoured by all
methods.}
\label{fig:condmaps}
\end{figure*}

\begin{figure*}[t]
\centering
\includegraphics[width=0.86\textwidth]{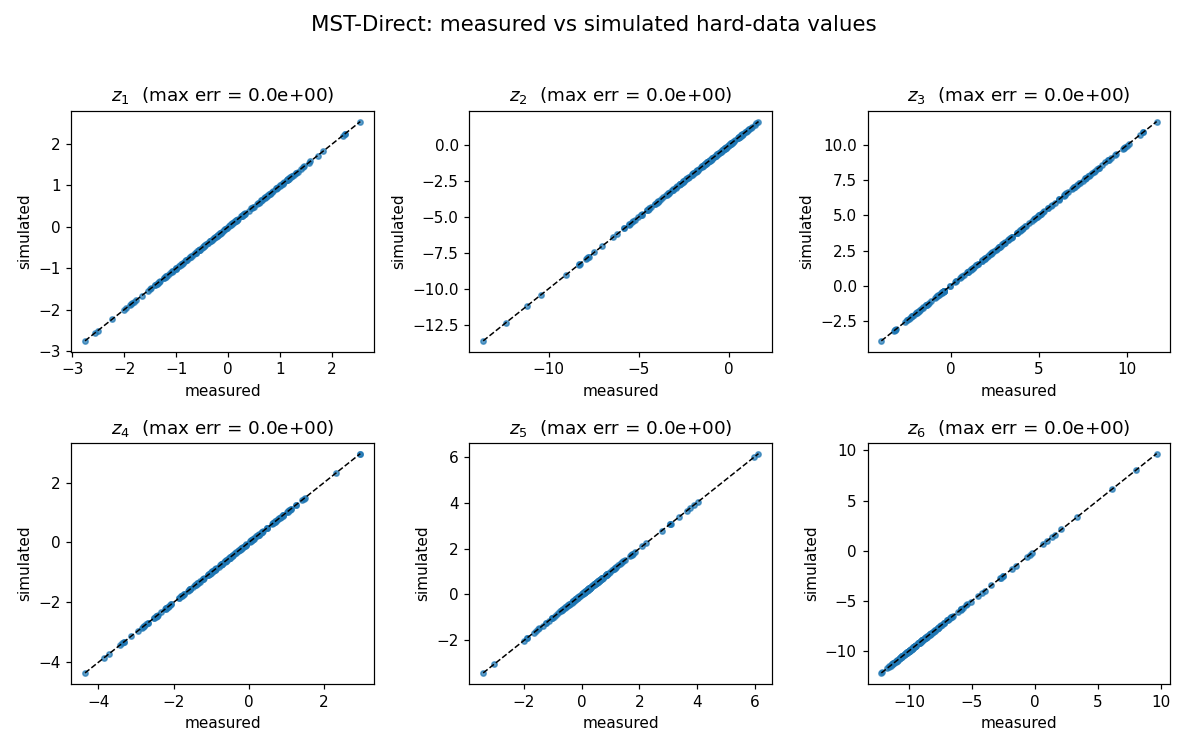}
\caption{Conditional MST-Direct: measured vs.\ simulated values at the 200
hard-data locations. All points lie on the $45^\circ$ line (maximum absolute
error $0$).}
\label{fig:condhonor}
\end{figure*}

\begin{figure*}[t]
\centering
\includegraphics[width=0.86\textwidth]{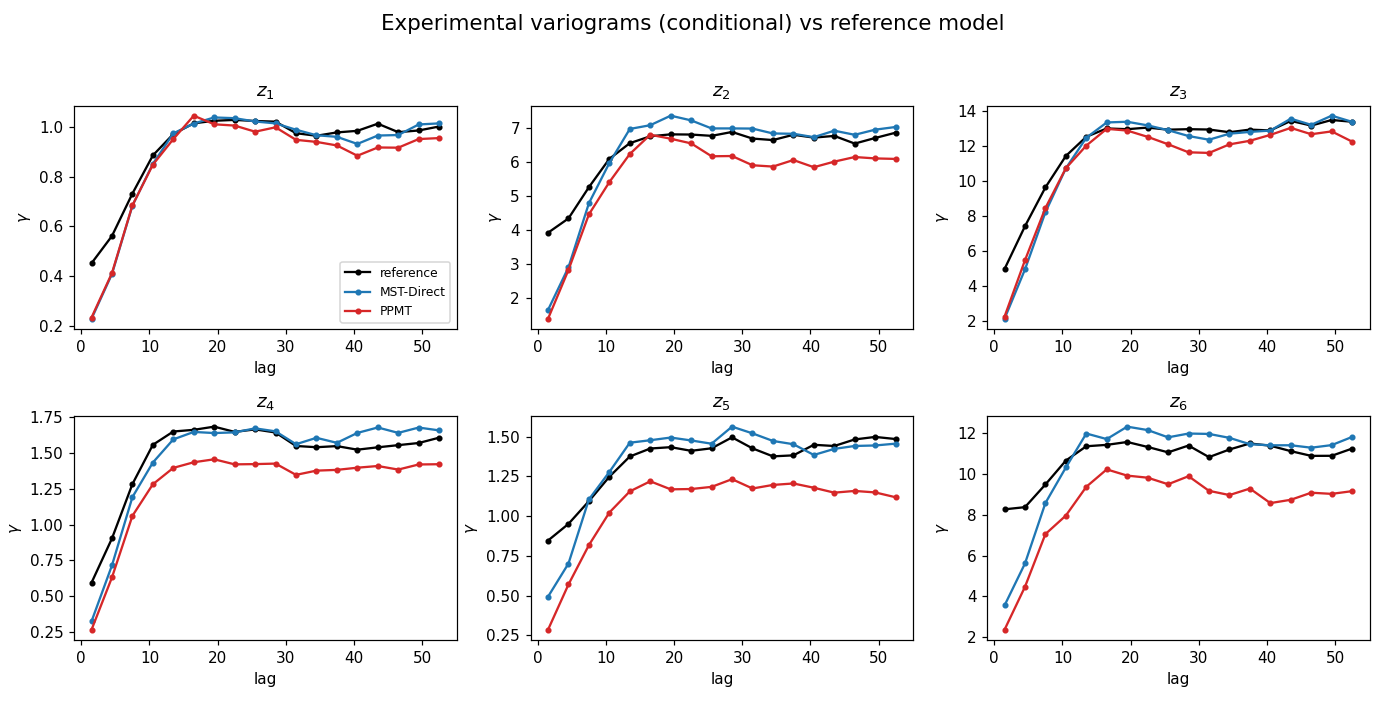}
\caption{Conditional experimental variograms: reference model (black),
MST-Direct (blue), PPMT (red).}
\label{fig:condvario}
\end{figure*}

For the conditional experiment, a reference model is built as an unconditional
realization on the $100\times100$ grid, and 200 of its nodes become hard data. Both methods honour the data ---
MST-Direct exactly by pinning (maximum absolute error $0$ over the 200 locations
and all six variables), PPMT to machine precision ($\sim10^{-14}$).
Fig.~\ref{fig:cond} compares the distributions (again MST-Direct exact, PPMT
approximate), with the MSE in Table~\ref{tab:cond}. Fig.~\ref{fig:condmaps}
shows the three realizations as maps (reference / PPMT / MST-Direct), all
honouring the data; Fig.~\ref{fig:condhonor} verifies the exact data
reproduction (measured vs.\ simulated on the $45^\circ$ line).
Fig.~\ref{fig:condvario} shows that both methods approximately reproduce the
reference variograms, MST-Direct tracking the sill exactly and PPMT slightly
undershooting it --- consistent with \cite{deFigueiredo2021dms}.

\begin{table*}[t]
\centering
\caption{Conditional histogram MSE vs.\ the reference ($\times10^{-5}$):
MST-Direct (left) and PPMT (right).}
\label{tab:cond}
\footnotesize
\begin{tabular}{lcccccc}
\multicolumn{7}{c}{\textbf{MST-Direct}}\\
\toprule
 & $z_1$ & $z_2$ & $z_3$ & $z_4$ & $z_5$ & $z_6$\\
\midrule
$z_1$ & 0 & 0 & 0 & 0 & 0 & 0\\
$z_2$ & 0 & 0 & 0 & 0 & 0 & 0\\
$z_3$ & 0 & 0 & 0 & 0 & 0 & 0\\
$z_4$ & 0 & 0 & 0 & 0 & 0 & 0\\
$z_5$ & 0 & 0 & 0 & 0 & 0 & 0\\
$z_6$ & 0 & 0 & 0 & 0 & 0 & 0\\
\bottomrule
\end{tabular}
\hspace{2em}
\begin{tabular}{lcccccc}
\multicolumn{7}{c}{\textbf{PPMT}}\\
\toprule
 & $z_1$ & $z_2$ & $z_3$ & $z_4$ & $z_5$ & $z_6$\\
\midrule
$z_1$ & 1.99 & 0.02 & 0.01 & 0.01 & 0.02 & 0.02\\
$z_2$ & 0.02 & 0.67 & 0.01 & 0.01 & 0.02 & 0.02\\
$z_3$ & 0.01 & 0.01 & 0.86 & 0.01 & 0.01 & 0.01\\
$z_4$ & 0.01 & 0.01 & 0.01 & 0.86 & 0.01 & 0.01\\
$z_5$ & 0.02 & 0.02 & 0.01 & 0.01 & 1.60 & 0.02\\
$z_6$ & 0.02 & 0.02 & 0.01 & 0.01 & 0.02 & 2.15\\
\bottomrule
\end{tabular}
\end{table*}

\section{Discussion}
The experiments close the three limitations left open in
\cite{Schmitz2026mst}. \emph{Scalability:} the sparse candidate-restricted
Sinkhorn runs the $40\,000$- and $10\,000$-node grids in well under a minute,
where the dense matcher would require a $40\,000^2$ coupling. \emph{Many
variables:} matching the target cloud onto an independent FFT-MA backbone
extends the method to six variables (and, in principle, more) without the
factorization error that grows with dimensionality in SCT/PPMT/DMS.
\emph{Conditioning:} pinning combined with a kriged backbone honours hard data
exactly while reproducing the spatial model. As in DMS, neither method controls
the cross-variograms directly, yet both reproduce the target statistics. The
main residual artifact is a short-lag component in the unconditional variograms,
controllable through the number of candidates and relational passes; anisotropic
variograms remain future work.

\section{Conclusion}
We extended MST-Direct to multivariate, conditional, large-grid geostatistical
simulation. The method matches the target value tuples onto an FFT-MA Gaussian
backbone through a scalable Sinkhorn optimal-transport step and honours hard
data by pinning, so the multivariate joint distribution is preserved exactly
while the prescribed spatial correlation is reproduced. On the 6-variate,
heteroscedastic, non-linear benchmark of Direct Multivariate Simulation, in both
unconditional and conditional settings, MST-Direct reproduces the target
distribution with zero histogram error and honours the conditioning data
exactly, outperforming PPMT on the joint distribution while matching it on the
spatial statistics. The approach is well suited to applications with complex,
strongly non-linear multivariate dependencies.

\section*{Code availability}
An open-source implementation (v2.0.0), together with the scripts that reproduce
the experiments in this paper, is released as the \texttt{mst-direct} Python
package~\cite{mstdirect_software}.

\section*{Acknowledgment}
The author thanks PX.Center for supporting this research.

\bibliographystyle{IEEEtran}
\bibliography{references}

\end{document}